\documentclass[lettersize,journal]{IEEEtran}
\usepackage{amsmath,amsfonts, amssymb}
\usepackage{array}
\usepackage[caption=false,font=normalsize,labelfont=sf,textfont=sf]{subfig}
\usepackage{textcomp}
\usepackage{stfloats}
\usepackage{url}
\usepackage{verbatim}
\usepackage{graphicx}
\usepackage{cite}
\usepackage{tikz}
\usepackage{xcolor}
\usepackage{algorithm}
\usepackage{algpseudocode}
\usepackage{multirow}
\usepackage{makecell}

\usetikzlibrary{
	positioning,
	arrows.meta,
	fit,
	shapes.geometric,
	calc, 
	decorations.markings
}

\renewcommand*{\arraystretch}{1.1}
\begin{document}
	
	\title{Denoising Diffusion Generative Models Secretly Calculate Attentions}
	
	\author{
		Farzan Haddadi\textsuperscript{$\dagger$}, 
		Leila Monfared\textsuperscript{$\dagger$}, 
		Ebrahim Rezaii\textsuperscript{$\dagger$}, 
		Mohammadreza Malek-Mohammadi\textsuperscript{*}, 
		Pejman Zakalvand\textsuperscript{$\dagger$}, 
		Narges Mokhtari\textsuperscript{$\dagger$}
		\thanks{\textsuperscript{$\dagger$} School of Electrical Engineering, Iran University of Science \& Technology, Tehran, IRAN}
		\thanks{\textsuperscript{*} Independent researcher}}

	\maketitle
	
	\begin{abstract}
		Denoising diffusion models are the dominant architecture for image generation, whereas most natural language generation and modeling are primarily handled by well-known transformer architectures employing attention mechanism. Here, we show that diffusion models also inherently use an attention mechanism very similar to that of transformers. Therefore, attention emerges as a universal machine learning principle, based on a general training objective. We also show similarities in basic functional principle of auto-encoders and attention-based models. These equivalences allows us to interchange these designs based on practical requirements. As an example, we can reformulate the diffusion framework to reduce the lengthy training process and computation-intensive image generation. Using this approach, a simplified algorithm is proposed for image generation which is based on attention mechanism. Results show that the attention-based implementation achieves comparable performance with significantly less effort and computational resources. 
	\end{abstract}
	
	
	\section{Introduction}
	\label{sec:intro}
	\IEEEPARstart{A}{rtificial} intelligence (AI) encompasses a multitude of proposed architectures for different tasks. Throughout the history of AI, it was a surprise that gradually efficient unified system architectures replaced this multitude of systems for various applications. An important turning point was the advent of the dot-product-based attention mechanism \cite{vaswany}. Transformers were first proposed for tasks of natural language processing (NLP) such as question-answering and chatbots \cite{llm}. Soon, it was realized that transformers also have stellar performance in vision applications with the advent of the vision transformer (ViT) model \cite{vit}. 
	
	In image generation, diffusion systems are arguably the best option \cite{ddpm}. Yet, for the best performance they are used in tandem with transformer-based components. Specifically, diffusion principle is used to determine the loss function and high-level train and test schemes, while in network level, an auto-encoder is trained which in turn uses attention mechanism and transformer architecture as its main component \cite{dit}.
	
	Here, we show that the diffusion principle is equivalent to the attention mechanism. Diffusion denoising probabilistic generative model (DDPM) \cite{ddpm}, trains a neural network to reconstruct the original image from its noisy copies. Then, starting from pure noise, the system can generate a realistic image based on the learned distribution of the dataset. 
	
	Detecting the presence of a signal from its noisy version is a classic statistical problem \cite{detection}. Specifically, when the noise is Gaussian, the solution is an inner-product detector. While in the context of machine learning theory, we train a neural network to solve this problem, we will show that this problem enjoys a closed-form solution which is basically the attention mechanism. This not only shows that the diffusion system is identical to the attention mechanism, but also provides a previously known statistical basis for the very attention mechanism as the closed form solution to the signal recovery problem in the Gaussian noise \cite{henderson}.
	
	The closed form solution of the diffusion system was first derived in \cite{analytic}. The authors calculate optimum score function of a diffusion system for a finite dataset. However, they did not notice the relation of their result to the well-known attention mechanism. Then \cite{analytic} extends the result to include additional structures like CNN network which is used in the DDPM implementation and develops a creativity theory for CNN diffusion systems based on locality and positional invariance properties. Here we use a different approach based on the diffusion loss function and convex optimization to reach the same closed form solution while extending the analysis to include resemblance to transformer and autoencoders. The equivalence between diffusion and transformer principles have broad theoretical and practical consequences in machine learning theory.
	
	It is not sufficient to understand the core diffusion principle in the forms of what we investigate as closed-form one-step solution or convergence trajectory. In practice, diffusion systems are implemented in the inner latent space of a Variational Auto-Encoder (VAE) \cite{stable}. Also, in practice the main neural network that performs denoising in the diffusion system is an AE. We argue that the same equivalence exists between attention and AE, in that both systems eventually project a random input on a learned data manifold \cite{manifold, ae_geometry}. Therefore, we reach to an equivalence between three of the most important modern AI systems, namely: ‌attention, diffusion , and AE. This enables us to interchange these systems depending on the situation. 
	
	DDPM diffusion system suffers from heavy computations in both training and generation which requires running an encoder hundreds of times on a random input. Some efforts have been made to reduce computational complexity specially in generation. A trend in the literatur is to reduce the steps for image generation using non-Markovian statistical models as in Diffusion Denoising Implicit Model (DDIM) \cite{ddim}. 
	
	Regarding the taining, DDIM is equivalent to DDPM and very computationally complex. It requires adding noise for hundreds of times to each image in dataset to teach network to denoise images with varying degree of noise effectively. Therefore, reducing computational complexity in the training phase is highly important. Here, we calculate a closed form denoiser which reduces the need for training a network on the dataset. 
	
	As a result of the equivalence of diffusion and attention, we conclude that deliberate interpolation between neighbor real images in the latent space results is an approximately realistic image. This interpolation can be refined and projected on the real image manifold by a limited recursive application of the outer AE. Overall, the proposed system can generate a realistic image without training an inner AE network, and just by using the dataset images. The results show good quality with considerably lower computations. 
	
	The manuscript is organized as follows:‌ The main result and discussion is presented in Sec. \ref{sec:attention}, where a background review, discussion on equivalence between diffusion and attention, and convergence topics are covered. AutoEncoder equivalence with attention and projection on data manifold are addressed in Sec. \ref{sec:autoencoder}. A data structure for fast attention calculation is discussed in Sec. \ref{sec:tree}. The proposed low complexity generative network is demonstrated in Sec. \ref{sec:algorithm}, while simulation results are exhibited in Sec. \ref{sec:simulation}.

	\section{Diffusion Systems Calculate Attention}
	\label{sec:attention}

	\subsection{Background}
	
	A diffusion system is trained to denoise noisy images in a recurssive manner. In the forward process, Gaussian noise is gradually added to the true image $x_0$ in hundreds or thousands of steps \cite{luo}: 
	\begin{align}
		\nonumber &‌  x_t = \sqrt{\alpha} \; x_{t-1} +‌\sqrt{1-\alpha} \; n_t \\
		\nonumber &   n_t \sim \mathcal{N} (0, I)
	\end{align}
	for $\alpha <‌1$.  Therefore, each image in the forward process can be written based on the true initial image as:
	\begin{equation}
		\label{eq:xtx0} x_t = \sqrt{\alpha^t} \; x_0 +‌\sqrt{1-\alpha^t} \; n'_t
	\end{equation}
	This eventually leads to a noise-only image at the final stage of the forward process since $\lim_{t\to+\infty} \alpha^t = 0$. We train a neural network to recover the true image $x_0$ from each noisy image $x_t$. Theoretically, this results in a neural network that can produce a realistic image from a pure noise and based on the true distribution of the data which is represented by the dataset used for training.

	\subsection{Prior Art}
	\label{sec:prior}
	
	In practice, we train the denoiser neural network on a finite dataset $\mathcal{D} = \{s_1, \ldots, s_N\}$. This finite dataset training situation was first analyzed in \cite{analytic}. Adding Gaussian noise incurs a mixture of Gaussian distribution on dataset points in signal space:
	
	\begin{equation}
		\label{eq:mixture}
		p_t(y) \; = \; \frac{1}{| \mathcal{D}|} \sum_{s \in \mathcal{D} } \mathcal{N}( y \; | \; \sqrt{\alpha^t} s \; , (1 - \alpha^t ) I ) 
	\end{equation}
	
	Then \cite{analytic} uses a flow matching continuous-time model which is solved backward in time for signal generation:
	
	\begin{equation}
		\label{eq:flow}
		-\Dot{y}_t = \gamma_t \left( y_t + s_t(y_t) \right) 
	\end{equation}
	in which $\gamma_t$ is a constant and $s_t(y) := \nabla_y \log p_t(y)$ is the score function. The score function of the mixture distribution in \eqref{eq:mixture} is then \cite{analytic}:  
	
	\begin{equation}
		\label{eq:score}
		s_t(y) = \frac{1}{1 - \alpha^t}  \sum_{s \in \mathcal{D}}  ( \sqrt{\alpha^t} s - y ) W_t(y,s) 
	\end{equation}
	\begin{equation}
		\label{eq:analytic}
		W_t(y,s) =  \frac{\mathcal{N}( y \; | \; \sqrt{\alpha^t} s \; , (1 - \alpha^t ) I )}{\sum_{s' \in \mathcal{D} } \mathcal{N}( y \; | \; \sqrt{\alpha^t} s' \; , (1 - \alpha^t ) I ) } 
	\end{equation}
	
	Then \cite{analytic} continues the analysis to develop a theory for combinatorial creativity of diffusion models when the denoiser is a CNN network.

	\subsection{Diffusion is Attention}
	\label{sec:main}
	
	Our analysis starts from the same finite dataset training assumption and then proceeds in a different direction to find the relation of the diffusion model to transformers and autoencoders as current dominant models in machine learning research.
	
	When we train the network on the finite dataset $\mathcal{D}$, the network learns to recover each $s_i$ from its noisy versions. We guess that when sampling the system for a new signal, although we use a random noise as the input, the system still outputs just an interpolation of the signals it had seen in the dataset since it has not seen anything else. 
	
	For signal $s_i$ and Gaussian noise $n$, DDPM minimizes Mean-Squared Error (MSE) loss function:
	\begin{equation}
		\label{eq:ddpm} \text{DDPM} : \quad \min_\theta \| \hat{x}_0(y,t; \theta) - s_i \|^2  \quad :‌ \quad 
		y = s_i + ‌n
	\end{equation}
	in which $x_0$ is the real signal, $\hat{x}_0$ its estimate, $t \in \{0, \ldots , T\}$ time index, and $\theta$ is the neural network's parameters. If we infinitely train the network with fixed $s_i$ and $y$, it learns to convert input $y$ to output $s_i$.
	
\begin{figure}
		\centering
		\begin{tikzpicture}[x=1mm,y=1mm]
			
			\draw (0,0) rectangle (80,50);
			
			\fill (30,10) circle (1pt);
			\fill (15,25) circle (1pt);
			\fill (60,25) circle (1pt);
			\fill (29.5,40) circle (1pt);
			
			\draw (30,10) -- (15,25);
			\draw (15,25) -- (60,25);
			\draw (30,10) -- (60,25);
			
			\node at (12,25) {$s_1$};
			\node at (28,7) {$s_2$};
			\node at (63,25) {$s_3$};
			\node at (29.5,43) {$y$};
			
			\draw (30,10) ++(0:9) arc (0:360:9);
			\draw (30,10) ++(60:30) arc (60:120:30);
			\draw (15,25) ++(0:10) arc (0:360:10);
			\draw (15,25) ++(10:21) arc (10:120:21);
			\draw (60,25) ++(0:10) arc (0:360:10);
			\draw (60,25) ++(140:34) arc (140:170:34);
			
		\end{tikzpicture}
		
		\caption{Geometry of diffusion system. Three signals $s_1$, $s_2$, and $s_3$ are used for system training. We are interested in calculating the output for random input $y$. We show that the output is the attention combination of training signals.}
		\label{fig:gauss}
	\end{figure}
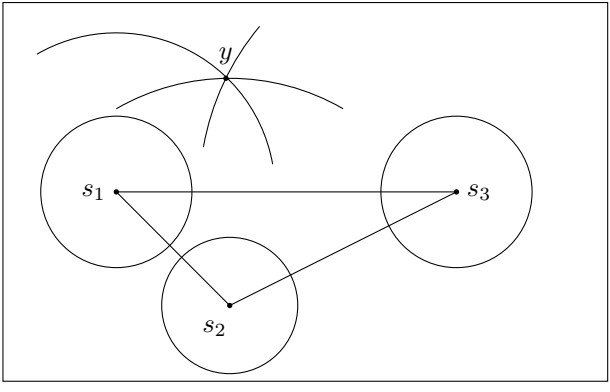

	But we do not train the network just for one sample image. Adding Gaussian noise to any signal $s_i$ for sufficiently many times, can reach any point $y$ in the signal space many times. A geometrical scheme of the problem is depicted in Fig. \ref{fig:gauss}. Reaching any fixed point $y$ from various signals $s_i$ occurs in accordance with the Gaussian distribution of the noise term $n$. Therefore, if we fix $y$, the system is trained with an average loss function:
	\begin{align}
		\label{eq:avloss} \min_\theta \quad \sum_{i=1}^N p(y|s_i) \| \hat{x}_0(y,t; \theta) - s_i \|^2 \\
		\nonumber p(y|s_i) = \frac{1}{\sqrt{2\pi} \sigma} \exp \left\{- \frac{\| y-s_i \|^2}{2\sigma^2} \right\}
	\end{align}

	For fixed $y$, \eqref{eq:avloss} is a Quadratic Programming (QP) convex optimization problem:
	\begin{align}
		\min_z \quad & \sum_{i=1}^N p_i \: \| z - s_i \|^2 \label{eq:socp}  \\
		& p_i \geqslant 0  \quad :‌ \quad i = 1, \cdots , N \label{eq:cone}
	\end{align}
	Where \eqref{eq:cone} is a conic criterion for the strictly convex optimization in \eqref{eq:socp}. The optimal solution for the above optimization problem can be derived using gradients \cite{boyd}:
	\begin{equation}
		\label{derivation} 
		\nabla_z \sum_{i} p_i \|z - s_i\|^2 
		= 2\sum_i p_i z - 2 \sum_i p_i s_i = 0
	\end{equation}
	\begin{equation}
		\label{eq:z*} z^* = \frac{\sum_{i=1}^N p_i s_i }{\sum_{i=1}^N p_i } = \frac{\sum_{i=1}^N \exp \left\{- \frac{\| y-s_i \|^2}{2\sigma^2} \right\} s_i }{\sum_{i=1}^N \exp \left\{- \frac{\| y-s_i \|^2}{2\sigma^2} \right\} }
	\end{equation}
	
	Assume that $\sigma$ is sufficiently small. Then just those signals $s_i$ that are in the neighborhood of $y$ are effective in the convex combination in \eqref{eq:z*}. Therefore, if $N$ is large enough and the space is sufficiently populated with data points, we can safely assume that $\|s_i\|$'s are approximately constant. Other situations also may hold the condition of approximately constant $\| s_i \|$ like in image dataset when total energy of most of the images are in the same range. In this case we will have:
	\begin{equation}
		\label{eq:att} z^* \simeq \frac{\sum_{i=1}^N \exp \left\{ \frac{< y , s_i > }{\sigma^2} \right\} s_i }{\sum_{i=1}^N \exp \left\{ \frac{ <‌y , s_i > }{\sigma^2} \right\} } = \text{Attention} \{ y , \{s_i\} \}
	\end{equation}
	in which $< \cdot \, , \cdot >$ denotes the inner vector product. 
	
	As we can see in \eqref{eq:att}, the optimum solution to the input signal $y$ in a diffusion system with fixed $\sigma$ is the attention vector. But this is just one step of the overall diffusion system since each step in the forward diffusion process has a distinct increasing value of $\sigma$. We will discuss this convergence issue in the next section \ref{sec:convergence}. 
	
	The derivation resulting in \eqref{eq:att} is also a mathematical basis for the very attention system. It shows that attention is the optimum solution for recovering signals in Gaussian noise as it was shown earlier in \cite{henderson}. Note that in the conventional attention system \cite{vaswany}, the noise variance $\sigma^2 = \sqrt{p}$, where $p$ is the dimension of the space. 
	
	This derivation also shows a basis for the softmax layer. The exponential form in \eqref{eq:att} with $\sigma = 1$, is the softmax output when $s_i$ is a standard one-hot label vector $\mathbb{I}_i$:
	\begin{align}
		&‌  z^*_i= \text{Softmax}_i \{ y \} = \frac{e^{y_i}}{\sum_{j=1}^N e^{y_j} }  \\
		&   z^* = \frac{\sum_{i=1}^N e^{< y , \mathbb{I}_i >} \mathbb{I}_i }{\sum_{i=1}^N e^{< y , \mathbb{I}_i >} } = \text{Attention} \{ y , \{\mathbb{I}_i\} \}
	\end{align}
	which shows the equivalence of softmax and attention layers:
	\begin{equation}
		\text{Softmax} \{ y \} = \text{Attention} \{ y , \{\mathbb{I}_i\} \}
	\end{equation}
	
	Note that the result in \eqref{eq:z*} can be deduced using the results in \cite{analytic}, replacing \eqref{eq:score} and \eqref{eq:analytic} in \eqref{eq:flow}. However, the authors in \cite{analytic} did not proceed to deduce the attention mechanism from their closed-form score function solution of the diffusion system in \eqref{eq:score}.

	\subsection{Convergence}
	\label{sec:convergence}
	In Sec. \ref{sec:main}, it was shown that each step of a diffusion system calculates attention for a predefined $\sigma$ value. The reverse generative process in DDPM starts with a random Gaussian $y$ and refines it with denoising network trained with noisy images from dataset. Theoretically, there should be hundreds of denoising networks each trained for denoising with a specific $\sigma$. But usually a shared network is trained and used for every values of $\sigma$ to reduce complexity. Therefore, output of each step is then recursively fed in the network to generate another less noisy output image. This is repeated hundreds of times to reach a high quality image. 
	
	\begin{figure}[!t]
		\centering
		\includegraphics[width=3in]{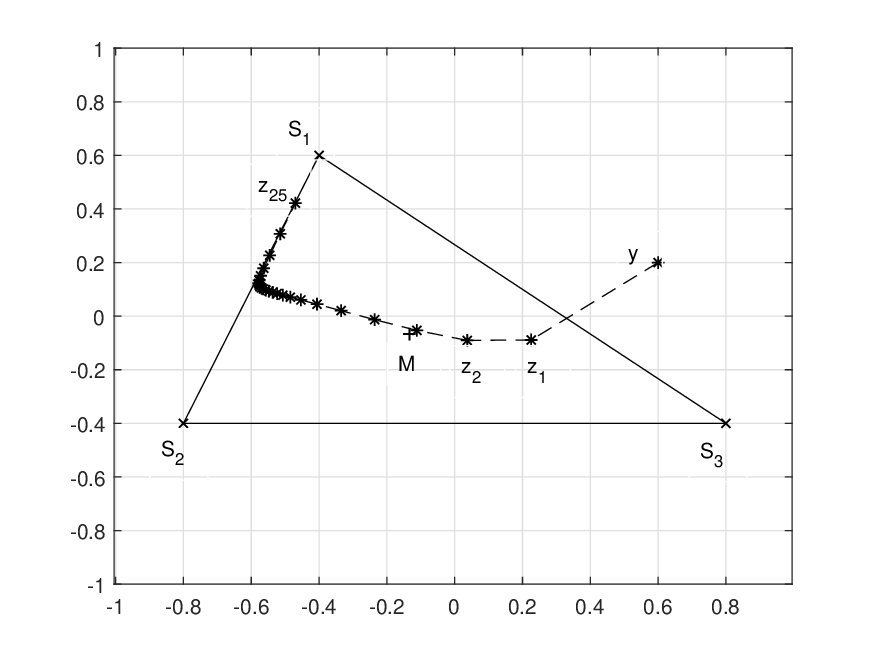}
		\caption{Convergence of a random point $y$ to a sparse solution through sequential attention calculation when $\sigma$ is decreasing. $M$ is the mean of the $s_i$ signals. Starting point $y$ passes near $M$ when $\sigma$ is large in the first stages.}
		\label{fig:convergence}
	\end{figure}
	
	As can be seen from \eqref{eq:xtx0}, $\sigma^2_t = 1 - \alpha^t$ is increasing with $t$ in the forward process. This will result in a decreasing $\sigma_t$ in the backward generative process. We start with a Gaussian random $y$. In the early stages when $\sigma$ is large, differences in distances $\| y-s_i \|$ in \eqref{eq:z*} make negligible effect. Therefore, the output sequence converges to the mean value of the nearby vectors. 
	
	When $\sigma$ gets small, differences in distances are promoted in \eqref{eq:z*} and the results quickly converge to the nearest signal, with some contribution from nearby signals. This is a sparse representation, as the simulation of \eqref{eq:z*} for a simple setup shows in Fig. \ref{fig:convergence}. In this simulation, $\sigma_t = 0.96 \, \sigma_{t+1}$ is a decreasing sequence which is decreased gradually from $1$ to $0.37$ in the reverse process. Although, as we will discuss later in Sec. \ref{sec:finite}, it seems that $\sigma$ does not converge to zero because of practical limitations in the denoising neural network design and training.
	
	The diffusion process initially converges to the mean point of the distribution. Then in the long term, it converges to the vicinity of some dense region of the dataset. This is in accordance with the fact that the diffusion process simulates the real data distribution as reflected in the dataset. 
	
	Knowledge about the convergence of the diffusion process enables us to circumvent the whole convergence process by directly using a convex combination of some nearby dataset signals as output of the generation process. This basically reduces not only the lengthy generation process, but also the very lengthy training process of denoising neural network.

	\subsection{Finite Training}
	\label{sec:finite}
	In this manuscript, we have generally assumed infinite training of denoising network. This enables the training process to cover every $y$ point in the space. It also enables the network through many gradient descent updates, to achieve the ground-truth closed form solution of the training process optimization. 
	
	This is rarely the case in practice. Due to limited resources like silicon, time, and energy, we usually perform minimal training up to the point of acceptable network performance. Here we discuss the effect of non-asymptotic training on the network behavior.
	
	First, the network does not experience every $y$ in the space during the training. It only sees finite adjacent inputs $\{ y_1 , \ldots, y_k\}$. It learns the correct direction of move from these inputs to outputs $\{o_1, \ldots, o_k\}$, respectively. Any new $y$ is sorrounded by multiple seen points. Since the network is a continouos system, we expect an average behavior in $y$ if the training set is sufficiently dense. This guides $y$ to regions of higher density which are experienced more in training and gradually we experience better refinements. 
	\begin{align}
		& o_i = \mathbb{D} \left( \mathbb{E} \left( y_i \right) \right) \\
		& y = \sum_{i=1}^k \alpha_i y_i  \quad  \to \quad  \mathbb{D} \left( \mathbb{E} \left( y \right) \right) = \sum_{i=1}^k \beta_i  o_i
	\end{align}
	where $ \{ \alpha_i\}$ and $ \{ \beta_i\}$ are two convex coefficient sets, $\mathbb{E}$ is the encoder function and $\mathbb{D}$ is the decoder.
	
	The second consequence of finite training is that the network does not learn the optimal solution of the optimization in \eqref{eq:avloss} for each training input $y_i$. We can model this training error as an additive noise in the output of the network. The same is true for the unseen input error and we can model the error in $\mathbb{D} \left( \mathbb{E} \left( y_i \right) \right)$ also as an additive noise.
	
	When noise is added to the output of the denoiser network in a diffusion system, the output will no longer be a sparse attention combination even when $\sigma$ of the diffusion system is very small. This is in fact a new additive noise in the backward process:
	\begin{equation}
		z_{t-1} = z_t + n''_t  \quad \Rightarrow \quad x_t = \sqrt{\alpha^t} x_0 + \sqrt{1-\alpha^t} \, n'_t - n''_t
	\end{equation}
	where $n''_t$ is the new noise term indicating errors emanating from finite training process. 
	
	Therefore, the overall effect of finite training can be modeled as a higher value of $\sigma$ which increases the number of intermittent signals. This means that although in the diffusion framework, theoretically $\sigma$ gradually converges to zero, practical limitations leads to a stable higher limiting value for $\sigma$. This prevents the optimization process to achieve a very sparse solution and the final output will be a convex combination of many signals.

	\section{Auto Encoders}
	\label{sec:autoencoder}
	
	In practice, diffusion systems use  auto-encoders as both the outer encoder layer that the diffusion system operates inside \cite{stable}, and the inner recurrent denoising network. Therefore, it is also crucial to understand working principles of AE to reduce computational complexity of the diffusion systems. In this section we use a different perspective to explain AE and attention system similarity.

	\subsection{Auto-encoders are attentions}
	
	As we showed in Sec. \ref{sec:attention}, there is a deep relationship between diffusion generative systems and attention mechanism. Here, we also show a strong resemblance between AEs, attention systems, and diffusion systems. This will present a great unity between three of the most important artificial intelligence systems currently in use.
	
	First of all, notice that the derivations in Sec. \ref{sec:main} is not limited to a diffusion system. In fact, it is a widespread setting in unsupervised learning systems. Assume we have an AE network which as usual use MSE loss function to reconstruct input images or other signals in an unsupervised manner. Regardless of the AE's inner structure, we assume perfect training of the system. Most of the times, an augmentation technique of adding Gaussian noise to limited dataset signals is used for better training and performance. Now, we can see that the mathematical setup is identical to the optimization in \eqref{eq:avloss}. Therefore, the same solution as in \eqref{eq:z*} applies to the problem. This certifies that such an AE system is equivalent to an attention system in \eqref{eq:att} with fixed $\sigma$ in the infinite training setup. Attention parameter $\sigma$ in \eqref{eq:z*} can be tuned according to the structural parameters of the AE for a satisfactory equivalence.
	
	The equivalence between AE and attention is double-sided. It means that we can in principle replace current attention systems with AEs. A good example for such a replacement is the linear attention \cite{linear_att}, which is known to be a finite-state attention mechanism with compressed rather than full state. This is reminiscent of recursive auto-encoder structure \cite{eff_att}, especially when we consider the causal implementation of the linear attention.

	\subsection{Auto-encoders and Manifolds}
	\label{sec:ae_manifold}

	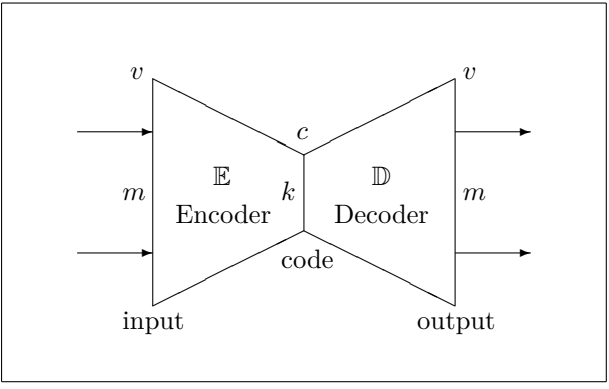
\begin{figure}
		\setlength{\unitlength}{1mm}
		\begin{picture}(80,50)
			
			\put(0,0){\line(1,0){80}}
			\put(0,0){\line(0,1){50}}
			\put(80,0){\line(0,1){50}}
			\put(0,50){\line(1,0){80}}
			
			\put(20,10){\line(0,1){30}}
			\put(40,20){\line(0,1){10}}
			\put(60,10){\line(0,1){30}}
			\put(20,10){\line(2,1){20}}
			\put(20,40){\line(2,-1){20}}
			\put(40,20){\line(2,-1){20}}
			\put(40,30){\line(2,1){20}}
			
			\put(16,24){$m$}
			\put(37,24){$k$}
			\put(61,24){$m$}
			\put(17,40){$v$}
			\put(61,40){$v$}
			\put(39,32){$c$}
			\put(16,7){input}
			\put(37,15){code}
			\put(55,7){output}
			\put(28,26){$\mathbb{E}$}
			\put(23,21){Encoder}
			\put(49,26){$\mathbb{D}$}
			\put(44,21){Decoder}
			
			\put(10,33){\vector(1,0){10}}
			\put(10,17){\vector(1,0){10}}
			\put(60,33){\vector(1,0){10}}
			\put(60,17){\vector(1,0){10}}
			
		\end{picture}	
		\caption{An auto-encoder with $m$-dimensional input/output vectors and $k$-dimensional code layer.}
		\label{fig:ae} 
	\end{figure}

	Assume an auto-encoder with $m$-dimensional input-output and $k$-dimensional code layer as in Fig. \ref{fig:ae}. Manifold hypothesis assumes that data from the dataset is located on a high-dimensional manifold in input space \cite{manifold}. It is well known that the autoencoder estimates the data manifold during the training phase \cite{ae_geometry}. We will show that AE projects the $m$-dimensional input signal on to a $k$-dimensional manifold which represents locus of data in the dataset and their meaningful augmentations in the space. 
	
	Assume that the AE is fully trained with the dataset and therefore, it can now regenerate signal $s$ in the output approximately. In the input space, we add a differential length vector $v_1$ to $s$. With high probability, no ReLU-like switch in the network is on its change position. Therefore, for any differential change in the input, the system is linear. Then, changing the input $s$ with $v_1$ results in a differential $c_1$ vector change in the code layer:
	\begin{equation}
		\mathbb{E} \left( s + v_1 \right) = \mathbb{E} (s) + c_1
	\end{equation}
	
	Continue these changes with orthogonal differential vectors $v_i$ that results each in code layer change of $c_i$. In the $k+1$'th step, the set of $\{c_1, \ldots, c_{k+1}\}$ is a linearly dependent set since the code layer vector space is $k$-dimensional. Therefore, a linear combination of these vectors sums to zero:
	\begin{equation}
		\sum_{i=1}^{k+1} \alpha_i c_i = 0
	\end{equation}
	
	The same set of $\alpha_i$ coefficients can be used to find a direction in the input space that keeps the code layer, and therefore the output constant:
	\begin{equation}
		\mathbb{D} \left( \mathbb{E} \left( s + \sum_{i=1}^{k+1} \alpha_i v_i \right) \right) = \mathbb{D} \left(  \mathbb{E} (s) +‌ \sum_{i=1}^{k+1} \alpha_i c_i \right) = s
	\end{equation}

	Therefore, we showed that any $k+1$ vectors $v_i$ in the input space can form a direction with null effect in the output. This direction is:
	\begin{equation}
		w = \sum_{i=1}^{k+1} \alpha_i v_i
	\end{equation}
	The local null space of the encoder at $s$ is defined as the local Affine vector space of all such $w$ vectors:
	\begin{equation}
		\mathbb{N}(s) := \left\{ \: ‌s+ w \; | \; \mathbb{E} (s+w) = \mathbb{E} (s) \; , \; \|w\| <‌\delta \right\}
	\end{equation}
	where $\delta$ is a small constant. The null space $\mathbb{N}(s)$ is local to $s$ and is effective just in a $\delta$-neighborhood of $s$. At most $k$ input  vectors $v_i$ can be chosen that does not lie in this local null space. Therefore:
	\begin{equation}
		\textrm{dim}(\mathbb{N}(s)) = m-k
	\end{equation}
	
	Suppose $\mathbb{M}(s)$ is the $k$-dimensional local Affine space orthogonal to the local null space:
	\begin{align}
		\nonumber \mathbb{M}(s) := \{ \: s+v \; | & ‌<v,w> = 0 \; :‌ \\
		& \forall w :‌ s+w \in \mathbb{N}(s) \; , \; \|v\| < \delta \, \}
	\end{align}

	In the vicinity of $s$, any displacement along the null space $\mathbb{N}(s)$ does not change the values of the code layer and therefore, the output. But displacement along manifold $\mathbb{M}(s)$ translates to changes in code layer and output. Therefore, the AE will tune its null space $\mathbb{N}(s_i)$ such that it direct farthest from the neighbors of $s_i$. This will set the estimated data subspace $\mathbb{M}(s_i)$ in the nearest position that best cover the neighbors of $s_i$ and orthogonal to the local null space. This is in accordance with autoencoder estimating a manifold with good fit to neighboring data points \cite{ae_geometry} with extensive training on sufficiently dense datasets.

	\begin{figure}[h!]
		\centering
		\begin{tikzpicture}[x=1mm,y=1mm]
			
			\draw (0,0) rectangle (80,50);
			
			\draw[thick,black]
			(5,5)
			.. controls (20,40) and (30,0) ..
			(42,30)
			.. controls (50,50) and (60,20) ..
			(75,30);
			
			\node[black] at (57,38) {$\mathbb{F}$};
			\node[black] at (28,16) {$s_i$};
			\node[black] at (30,43) {$y$};
			
			\draw[thick,black,-{Latex[length=3mm]}]
			(30,40)
			.. controls (31,31) ..
			(40,25.5);
			
			\node[black] at (30,40) {$\times$};
			
			\pgfmathsetseed{121}
			
			\foreach \i in {0,...,10} {
				\pgfmathsetmacro{\t}{(\i * 3 + rnd * 2)/31}
				
				\path 
				(5,5)
				.. controls (20,40) and (30,0) ..
				(42,30)
				coordinate[pos=\t] (p);
				
				\pgfmathsetmacro{\dx}{(rnd-0.5)*1.5}
				\pgfmathsetmacro{\dy}{(rnd-0.5)*1.5}
				
				\fill (p) circle (0.5);
			}
			
			\foreach \i in {0,...,9} {
				\pgfmathsetmacro{\t}{(\i * 3 + rnd * 2)/31}
				
				\path 
				(42,30)
				.. controls (50,50) and (60,20) ..
				(75,30)
				coordinate[pos=\t] (p);
				
				\pgfmathsetmacro{\dx}{(rnd-0.5)*1.5}
				\pgfmathsetmacro{\dy}{(rnd-0.5)*1.5}
				
				\fill (p) circle (0.5);
			}
			
		\end{tikzpicture}
		\caption{The dataset $\mathcal{D}$ is assumed dense enough that the data manifold $\mathbb{F}$ can be followed by the auto encoder after sufficient training. Input $y$ is projected onto the data manifold in a curved path following the local null spaces $\mathbb{N}$ and eventually arrives at $\mathbb{F}$ in a orthogonal direction.}
		\label{fig:densemanifold}
	\end{figure}
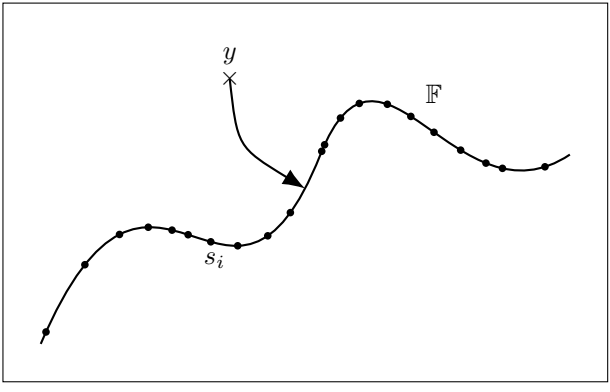

	Here we assume that the dataset is dense enough that the continuity of the data manifold can be tracked by the AE as in Fig.~\ref{fig:densemanifold} . We also assume that the AE does not have a specific structure (like convolutional or any other weight sharing scheme). This helps the AE to follow any direction in the data space without any bias. It is well-known that networks like convolutional networks better estimate the data manifold. This is because some of the data structure, like image content translation invariance is embedded in such networks. This is equivalent to more data points which helps the network better estimate the data manifold.

	Any input $y$ outside the manifold  defines a specific local null space $\mathbb{N}(y)$ which aggregates directions of moves that does not change the code layer and output layers values. Therefore, AE moves $y$ in a curved path along local null spaces to finally reach the AE estimated data manifold $\mathbb{F}$ in $h$. Since $\mathbb{N}(h) \perp \mathbb{M}(h)$ we reach to the AE manifold perpendicularly in the last touch as is shown in Fig.~\ref{fig:densemanifold}.

	\subsection{Attention and manifold}
	\label{sec:att_manifold}
	
	Here, we show that the attention mechanism also projects data on a low dimensional manifold. This is in-line with our core insight that basically these three structures:‌ AE, attention, and diffusion systems are similar. 
	
\begin{figure}[t!]
		\centering
		\begin{tikzpicture}[x=1mm,y=1mm]
			
			\draw (0,0) rectangle (80,45);
			
			\draw[thick]
			(10,5)
			.. controls (50,5) ..
			(70,30);
			
			\draw[thick]
			(10,20)
			.. controls (50,20) ..
			(70,40);
			
			\draw[dash dot, thick]
			(10,5) -- (9,7) -- (10,9) -- (9,11) -- (10,13)
			-- (9,15) -- (10,17) -- (9,19) -- (10,20);
			
			\draw[dash dot, thick]
			(70,30) -- (71,31.5) -- (70,33) -- (71,34.5)
			-- (70,36) -- (71,38) -- (70,40);
			
			\node at (57,22) {$\mathbb{F}$};
			\node at (18,17) {$s_{i_1}$};
			\node at (20,10) {$s_{i_2}$};
			\node at (40,10) {$s_{i_3}$};
			\node at (37,17) {$s_{i_4}$};
			\node at (29,33) {$y$};
			\node at (29,30) {$\times$};
			\node at (26.5,13) {$h$};
			\node at (29,13) {$*$};
			
			\draw[thick,-{Latex[length=3mm]}]
			(29,30) -- (29,13);
			
			\draw (21,17) -- (23,10);
			\draw (23,10) -- (36,10);
			\draw (36,10) -- (33,17);
			\draw (33,17) -- (21,17);
			
			\fill (23,10) circle (0.5);
			\fill (36,10) circle (0.5);
			\fill (21,17) circle (0.5);
			\fill (33,17) circle (0.5);
			
		\end{tikzpicture}
		\caption{Geometry of attention mechanism. Point $y$ with a small offset from the data manifold $\mathbb{F}$ is projected orthogonally to the Affine subspace spanned by nearby signals on the data manifold by attention mechanism.}
		\label{fig:att_manifold}
	\end{figure}
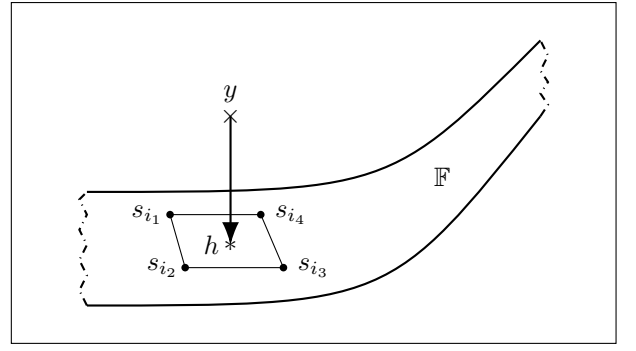

	Assume that $y$ is near the data manifold. Dataset satisfies conditions that we can use \eqref{eq:z*} in place of the better-known attention formula in \eqref{eq:att}.
	Also, assume that $\sigma$ is small enough that only $k$ nearest signals:
	\begin{equation}
		A_i := \{ s_{i_j} \}_{j=1}^k
	\end{equation}
	are effective in the attention sum in \eqref{eq:z*}. Then, the attention coefficients are:
	\begin{equation}
		\label{eq:att_proj} p_i \simeq \frac{\exp \left\{- \frac{\| y-s_i \|^2}{2\sigma^2} \right\} }{\sum_{s \in A_i} \exp \left\{- \frac{\| y-s \|^2}{2\sigma^2} \right\} } =: p_i (y,A_i)
	\end{equation}
	The adjacency set $A_i$ forms a $k-1$-dimensional Affine subspace. Assume that the orthogonal projection of $y$ onto this subspace is $h$ as in Fig. \ref{fig:att_manifold}. Then:
	\begin{equation}
		\| y-s_i \|^2 = \| y-h \|^2 + \| h-s_i \|^2
	\end{equation}
	Then $\| y-h \|^2$ is shared in numerator and denominator of \eqref{eq:att_proj}, eliminating it:
	\begin{equation}
		\label{eq:att_proj_h} p_i = \frac{\exp \left\{- \frac{\| h-s_i \|^2}{2\sigma^2} \right\} }{\sum_{s \in A_i} \exp \left\{- \frac{\| h-s \|^2}{2\sigma^2} \right\} } = p_i(h, A_i)
	\end{equation}
	Output of $h$ from the attention mechanism is a convex combination of $s \in A_i$ and therefore is on their Affine subspace. Also, the fact that the output of $y$ is the same as the output for $h$ shows that there is a local Affine null space orthogonal to the local affine space of nearby signals in $A_i$ that does not change the output of the attention.

	\subsection{Auto-encoders resemble attention}
	
	The attention mechanism when act on a point $y$ near the data manifold $\mathbb{F}$, orthogonally projects it on the manifold at point $h$, and then transform $h$ to another point in the convex hull of signals in $A_i$ in the output. Overall, we can conclude that the attention mechanism projects any point $y$ near the data manifold to some point on the manifold too and this is what also auto encoders do.
	
	The attention mechanism depends on the parameter $\sigma$, while the AE is characterized by its code layer dimension $k$. The equivalence between attention and AE requires fine tuning these parameters. As stated before, setting a small $\sigma$ is required in order that approximately $k+1$ nearby signals are effective in the attention. This will result in the estimation of a $k$-dimensional manifold for data similar to an AE with $k$-dimensional code layer. This is not an exact equivalence because the required $\sigma$ parameter may not be constant in different parts of the input space.
	
	In a conventional diffusion system, we have an outer autoencoder which compresses data to a lower dimension. Then a recursive inner autoencoder denoises an input noise. We have shown that not only the high level diffusion framework is equivalent to the attention mechanism, but also both autoencoders in the network structure realization perform in a similar way to the attention mechanism. Therefore, we can alternate between these three structures based on the situation.
	
	If attention and auto encoder structures are basically similar, what is their difference?‌ The difference lies in where the computational load is concentrated. In an auto encoder, we use vast amount of data in training phase while the test phase is as simple as one forward calculation of the network. The attention mechanism is reverse. Calculating attention does not require any training while all the computational load is concentrated in the test time where the forward pass of the network requires vast amount of product calculations with a long context of vectors. Therefore, we may be able to reduce the complex training phase of the auto encoder and diffusion systems by using some simpler form of the attention mechanism.

	\section{Tree Cross Attention}
	\label{sec:tree}
	
	The problem with the attention mechanism is that it involves calculating the dot product of the input $y$ with every image in the dataset. These image datasets are usually very large in industry-level network training. Since usually $\sigma$ parameter of attention is small, it suffices to calculate the attention just for nearby signals. But finding the nearbies requires calculating the distances to all images in the dataset which is equivalent to inner product computation. 
	
	Here, a data structure can lead to considerable reduction in the computational complexity. Calculating the dot product of $y$ with every images in the dataset is of $\mathcal{O}(|\mathcal{D}|)$ complexity. But since there is a great deal of redundancy in this calculation, we can save some information for reuse.

	Data can be order in a tree structure based on similarity \cite{algorithm}. Each node in the tree is an image and any new image is tested with every nodes in each tree level to see with which branch it is more similar. If there was a similar branch, the new image goes down to that branch's next level children. If there were no similar branch, make a new branch in the current level. 
	
	In this way, a balanced tree can be constructed in which similar and nearby images are clustered in a branch. Then for each new input signal $y$, we need a few inner product calculations of $\mathcal{O}(\log |\mathcal{D} |)$ which is the tree depth to find its neighbors for efficient attention calculations. This method is called tree cross attention and is presented in the literature \cite{tree_cross}, for the situations where base attention vectors are constant. Therefore, attention calculation is not restrictive in an image generation scenario.

\section{Proposed Generation Algorithm}
\label{sec:algorithm}

	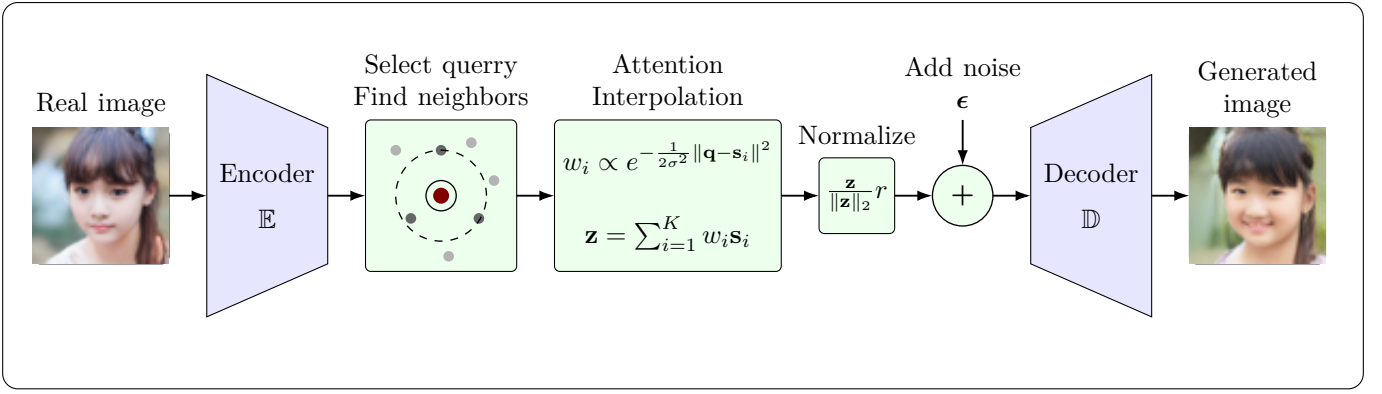
\begin{figure*}[!t]
		\centering

		\begin{tikzpicture}
			\node[
			draw,
			rounded corners=2mm,
			minimum width=18cm,
			minimum height=5.1cm,
			inner sep=5mm
			] {};
			
			\node[
			draw,
			rounded corners=1mm,
			inner sep=0pt
			] at (-7.7,0)
			{\includegraphics[
				width=18mm,
				height=18mm
				]{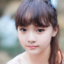}};
				
			\node[align=center] at (-7.7,1.2)
			{Real image};
			
			\draw[-{Latex[length=2mm,width=1.4mm]}, line width=0.7pt]
			(-6.8,0) -- (-6.3,0);
			
			\pgfdeclareshape{encoder shape}{
				\savedanchor\northwest{
					\pgfpoint{-12mm}{16mm}
				}
				\savedanchor\southeast{
					\pgfpoint{8mm}{-16mm}
				}
				
				\anchor{center}{\pgfpoint{0}{0}}
				\anchor{north}{\pgfpoint{0}{16mm}}
				\anchor{south}{\pgfpoint{0}{-16mm}}
				\anchor{west}{\pgfpoint{-12mm}{0}}
				\anchor{east}{\pgfpoint{8mm}{0}}
				
				\backgroundpath{
					\pgfpathmoveto{\pgfpoint{-12mm}{16mm}}
					\pgfpathlineto{\pgfpoint{4mm}{9mm}}
					\pgfpathlineto{\pgfpoint{4mm}{-9mm}}
					\pgfpathlineto{\pgfpoint{-12mm}{-16mm}}
					\pgfpathclose
				}
			}
			
			\node[
			encoder shape,
			draw,
			align=center,
			fill=blue!10
			] at (-5.1,0)
			{};
			
			\node[align=center] at (-5.5,0.3)
			{Encoder};
			
			\node[align=center] at (-5.5,-0.3)
			{$\mathbb{E}$};
			
			\draw[-{Latex[length=2mm,width=1.4mm]}, line width=0.7pt]
			(-4.7,0) -- (-4.2,0);

			\node[
			draw,
			rounded corners=2pt,
			minimum width=20mm,
			minimum height=20mm,
			inner sep=3mm,
			fill=green!7
			] (knn) at (-3.2,0) {};
			
			\fill[gray!60] ($(knn.center)+(-6mm, 6mm)$) circle (2pt);
			\fill[gray!60] ($(knn.center)+( 4mm, 7mm)$) circle (2pt);
			\fill[black!60] ($(knn.center)+(-4mm,-3mm)$) circle (2pt);
			\fill[black!60] ($(knn.center)+( 5mm,-3mm)$) circle (2pt);
			\fill[gray!60] ($(knn.center)+( 7mm, 2mm)$) circle (2pt);
			\fill[black!60] ($(knn.center)+(0mm, 6mm)$) circle (2pt);
			\fill[gray!60] ($(knn.center)+( 1mm,-8mm)$) circle (2pt);
			
			\fill[red!50!black]
			(knn.center) circle (3pt);	
			
			\draw[
			line width=0.5pt
			]
			(knn.center) circle (2mm);
			
			\node[align=center] at (-3.2,1.5)
			{Select querry \\ Find neighbors};
			
			\draw[
			dashed,
			line width=0.5pt
			]
			(knn.center) circle (6mm);
			
			\draw[-{Latex[length=2mm,width=1.4mm]}, line width=0.7pt]
			(-2.2,0) -- (-1.7,0);
			
			\node[
			draw,
			rounded corners=2pt,
			minimum width=30mm,
			minimum height=20mm,
			inner sep=3mm,
			fill=green!7
			] (knn) at (-0.2,0) {};
			
			\node[align=center] at (-0.2,1.5)
			{Attention \\ Interpolation};
			
			\node[align=center] at (-0.2,0.5)
			{$ w_i \propto e^{-\frac{1}{2\sigma^2} \| \mathbf{q} - \mathbf{s}_i \|^2}$};
			
			\node[align=center] at (-0.2,-0.5)
			{$ \mathbf{z} = \sum_{i=1}^K w_i \mathbf{s}_i$};
			
			\draw[-{Latex[length=2mm,width=1.4mm]}, line width=0.7pt]
			(1.3,0) -- (1.8,0);
			
			\node[
			draw,
			rounded corners=2pt,
			minimum width=10mm,
			minimum height=10mm,
			inner sep=3mm,
			fill=green!7
			] (knn) at (2.3,0) {};
			
			\node[align=center] at (2.3,0.8)
			{Normalize};
			
			\node[align=center] at (2.3,0)
			{$ \frac{\mathbf{z}}{\| \mathbf{z}\|_2} r$};
			
			\draw[-{Latex[length=2mm,width=1.4mm]}, line width=0.7pt]
			(2.8,0) -- (3.3,0);
			
			\draw[
			line width=0.5pt,
			fill=green!7
			]
			(3.7,0) circle (4mm);
			
			\node[align=center] at (3.7,0)
			{$\boldsymbol{+}$};	
			
			\draw[-{Latex[length=2mm,width=1.4mm]}, line width=0.7pt]
			(3.7,1) -- (3.7,0.4);
			
			\node[align=center] at (3.7,1.5)
			{Add noise \\ $\boldsymbol{\epsilon}$};
			
			\draw[-{Latex[length=2mm,width=1.4mm]}, line width=0.7pt]
			(4.1,0) -- (4.6,0);	
			
			\pgfdeclareshape{decoder shape}{
				\savedanchor\northwest{
					\pgfpoint{-12mm}{16mm}
				}
				\savedanchor\southeast{
					\pgfpoint{8mm}{-16mm}
				}
				
				\anchor{center}{\pgfpoint{0}{0}}
				\anchor{north}{\pgfpoint{0}{16mm}}
				\anchor{south}{\pgfpoint{0}{-16mm}}
				\anchor{west}{\pgfpoint{-12mm}{0}}
				\anchor{east}{\pgfpoint{8mm}{0}}
				
				\backgroundpath{
					\pgfpathmoveto{\pgfpoint{-12mm}{9mm}}
					\pgfpathlineto{\pgfpoint{4mm}{16mm}}
					\pgfpathlineto{\pgfpoint{4mm}{-16mm}}
					\pgfpathlineto{\pgfpoint{-12mm}{-9mm}}
					\pgfpathclose
				}
			}
			
			\node[
			decoder shape,
			draw,
			align=center,
			fill=blue!10
			] at (5.8,0)
			{};
			
			\node[align=center] at (5.4,0.3)
			{Decoder};
			
			\node[align=center] at (5.4,-0.3)
			{$\mathbb{D}$};
			
			\draw[-{Latex[length=2mm,width=1.4mm]}, line width=0.7pt]
			(6.2,0) -- (6.7,0);	
			
			\node[
			draw,
			rounded corners=1mm,
			inner sep=0pt
			] at (7.6,0)
			{\includegraphics[
				width=18mm,
				height=18mm
				]{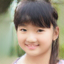}};
				
			\node[align=center] at (7.6,1.4)
			{Generated \\ image};
			
		\end{tikzpicture}
		\caption{Proposed attention-based generative architecture.}
		\label{fig:blockdiagram}
	\end{figure*}

	Theoretical equivalence of diffusion and attention systems helps to reduce the computational complexity of image generation. We have seen in Sec. \ref{sec:convergence} that any noise input to a diffusion system experiences a long road to its final position in latent space.
	
	Since the solution to diffusion is an approximately sparse one (Fig. \ref{fig:convergence}), we can start image synthesize from one of dataset images. Since attention to latent signals is approximately constant in the neighborhood of the image latent, we can calculate the attention of the latent signal to other neighbor signals. 
	
	This attention calculation helps the resulting signal to reside on or very near the data manifold since it is equivalent to passing the signal once through the hypothetical diffusion trained AE denoiser. Therefore, we can pass the resulting latent through the decoder and reach to a high quality innovative image. Since there may be slight disposition from the data manifold, we can pass the resulting image a few times through the AE to better fit it to the data manifold. 
	
	Therefore, there will be no need to train a denoiser auto encoder as the working engine of the diffusion system. We just need to train  an outer AE which projects the input images to the low-dimensional latent space. Then in the generation phase, we just rely on attention calculation near the signal latents. Therefore, we not only reduced the training phase calculations, but also in the test phase, we reduce the long journey of the input noise to the distribution mean value and then to nearby of one of the signals. We just perform the final step with calculating the attention of the signal to its neighbbors including itself. The block diagram of the proposed image generation algorithm is depicted in Fig. \ref{fig:blockdiagram}.
	
	The generation process is done in the learned latent space of trained VAE. As described in Alg. \ref{alg:sampling}, for a random latent query $\mathbf{q}$, we find its $K$ nearest neighbors $\{\mathbf{s}_k\}_{k=1}^{K}$. New latent representation is generated through attention interpolation between the query and its neighbors. The generated latent representation is then passed through the VAE decoder to obtain a new image. Since nearby latent representations correspond to semantically related data, their interpolation produce meaningful intermediate representations. 
	
\begin{figure}[t]
		\centering
		
		\begin{minipage}[t]{0.32\textwidth}
			\centering

			\setlength{\tabcolsep}{1pt}
			\setlength{\fboxrule}{1pt}
			\setlength{\fboxsep}{1pt}
			
			\begin{tabular}{@{}ccc@{}}
				{\small NN1} & {\small NN2} & {\small NN3} \\[1mm]
				
				\fbox{
					\includegraphics[width=0.28\linewidth]{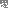}
				}
				&
				\fbox{
					\includegraphics[width=0.28\linewidth]{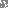}
				}
				&
				\fbox{
					\includegraphics[width=0.28\linewidth]{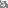}
				}\\[2mm]
			\end{tabular}

		\end{minipage}
		
		\caption{
			Spatial attention weight maps obtained for the top-$3$ nearest
			neighbors of the query on the FFHQ dataset. Each map illustrates
			the spatial contribution of the corresponding neighbor region to the
			generated latent representation.
		}
		\label{fig:spatial_attention}
	\end{figure}

	Although using plain attention between querry and its neighbor images produce meaningful images, we use a fine grain spatial attention mechanism to get better quality. Spatial attention is calculated based on the distance of each pixel depth vector in querry to its corresponding pixel depth vector in a neighbor image as presented in Alg. \ref{alg:spatial_attention}. This generates a spatial attention map that determines the contribution of each pixel to the generated sample. Example spatial attention maps in Fig. \ref{fig:spatial_attention} illustrate how the proposed method selectively exploits spatial information from neighbors during latent-space sampling. Note that spatial attention resembles the patch similarity selection mechanism that is developed as the theoretical solution of a convolutional diffusion network in \cite{analytic}.

	\begin{algorithm}[t]
		
		
		\caption{Attention-Based Image Generation}
		\label{alg:sampling}
		\begin{algorithmic}[1]
			
			\State \textbf{Input:} Latents
			$\{\mathbf{s}_i\}_{i=1}^{N}$,
			query $\mathbf{q}$,
			number of neighbors $K$,
			attention parameter $\sigma$,
			noise scale $\eta$
			
			\State \textbf{Output:} Generated image $\mathbf{x}$
			
			\State $r
			\gets
			\frac{1}{N}
			\sum_{i=1}^{N}
			\|\mathbf{s}_i\|_2$
			
			\State $d_i
			\gets
			\|\mathbf{q} - \mathbf{s}_i\|_2$
			\quad $\forall i$
			
			\State $\mathcal{S}
			\gets
			\{\mathbf{s}_i\mid i\in  \operatorname{TopK}(-d_i) \}$
			
			\State $\mathbf{W}
			\gets
			\operatorname{Spatial Attention}
			(\mathbf{q},\mathcal{S},\sigma)$
			
			\State $\mathbf{z}
			\gets
			\sum_{\mathbf{s}_i \in \mathcal{S}}
			\mathbf{W}_{i}
			\odot
			\mathbf{s}_{i}$
			
			\State $\mathbf{z}
			\gets
			r
			\frac{\mathbf{z}}
			{\| \mathbf{z}\|_2+\epsilon}$
			
			\State $\mathbf{z}
			\gets
			\mathbf{z}
			+
			\eta\boldsymbol{\epsilon},
			\quad
			\boldsymbol{\epsilon}
			\sim
			\mathcal{N}(\mathbf{0},\mathbf{I})$
			
			\State $\mathbf{z}_{q}
			\gets
			\operatorname{Quantize}(\mathbf{z})$
			
			\State $\mathbf{x}
			\gets
			\mathbb{D}(\mathbf{z}_{q})$
			
			\For{$j=1,\ldots,3$}
			\State $\mathbf{x}
			\gets
			\operatorname{AE}
			\left( \mathbf{x} \right)$
			\EndFor
			
			\State \Return
			$\mathbf{x}$
			
		\end{algorithmic}
	\end{algorithm}

	\begin{algorithm}[t]
		
		
		\caption{Spatial Attention}
		\label{alg:spatial_attention}
		\begin{algorithmic}[1]
			
			\State \textbf{Input:} Query $\mathbf{q} \in \mathbb{R}^{D \times H \times W}$, neighbor latents
			$\mathcal{S}=\{\mathbf{s}_i\}_{i=1}^{K}$, scale parameter $\sigma$
			
			\State \textbf{Output:} Attention weights
			$\mathbf{W}\in\mathbb{R}^{K\times H\times W}$
			
			\For{$i = 1,\ldots,K$}
			
			\State $\mathbf{s} 
			\gets
			\mathbf{s}_i$
			
			\ForAll{$j,k$}
			
			\State $d_{ijk}
			\gets
			\left\| \mathbf{q}_{:jk} - \mathbf{s}_{:jk}
			\right\|_2$
			
			\State $p_{ijk}
			\gets
			-\dfrac{d_{ijk}^2}
			{\sigma^2+\epsilon}$
			
			\State $w_{ijk}
			\gets
			\operatorname{softmax}_{i}(p_{ijk})$
			
			\EndFor
			
			\EndFor
			
			\State \Return $\mathbf{W}
			\gets
			\{w_{ijk}\}$
			
		\end{algorithmic}
	\end{algorithm}

	\section{Simulation Results}
	\label{sec:simulation}
	
	We train the model using a P100 GPU on the MNIST, Fashion-MNIST, FFHQ, and CelebA datasets with an image size of $28\times28$ for MNIST and Fashion-MNIST, $64\times64$ for FFHQ and $128\times128$ for CelebA.
	The Adam optimizer is used with
	$\beta_1=0.9$ and $\beta_2=0.99$,
	and a learning rate of $1\times10^{-4}$. For outer VAE training, we use a combination of $L_1$ loss, perceptual loss based on VGG network and an adversarial (GAN) loss of StyleGAN discriminator with the following architecture:
	
	\begin{equation}
		L_{\mathrm{rec}}= L_1 + L_{\mathrm{perceptual}} + L_{\mathrm{adv}}
	\end{equation}

	In our approach, generation is performed by local interpolation which is substantially faster than iterative reverse diffusion. The diffusion method requires $0.17$ second versus the proposed method just $0.02$ second to generate an image on the FFHQ dataset, hence a $8.5\times$ speedup. 
	
\begin{figure*}[t]
		\centering
		
		\begin{minipage}[t]{0.32\textwidth}
			\centering
			
			\rule{\linewidth}{0.4pt}\\[-1mm]
			{\small FFHQ}\\[-2mm]
			\rule{\linewidth}{0.4pt}\\[1mm]
			\setlength{\tabcolsep}{0.5pt}
			\begin{tabular}{@{}ccccc@{}}
				\includegraphics[width=0.19\linewidth]{2/q1.png} &
				\includegraphics[width=0.19\linewidth]{2/s1.png} &
				\includegraphics[width=0.19\linewidth]{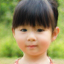} &
				\includegraphics[width=0.19\linewidth]{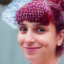} &
				\includegraphics[width=0.19\linewidth]{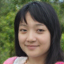} \\[-1mm]
				
				\includegraphics[width=0.19\linewidth]{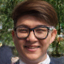} &
				\includegraphics[width=0.19\linewidth]{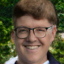} &
				\includegraphics[width=0.19\linewidth]{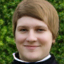} &
				\includegraphics[width=0.19\linewidth]{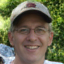} &
				\includegraphics[width=0.19\linewidth]{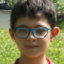} \\[-1mm]
				
				\includegraphics[width=0.19\linewidth]{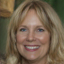} &
				\includegraphics[width=0.19\linewidth]{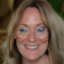} &
				\includegraphics[width=0.19\linewidth]{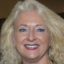} &
				\includegraphics[width=0.19\linewidth]{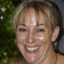} &
				\includegraphics[width=0.19\linewidth]{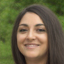}
			\end{tabular}
			
			\rule{\linewidth}{0.4pt}
			
		\end{minipage}
		\hfill
		\begin{minipage}[t]{0.32\textwidth}
			\centering
			
			\rule{\linewidth}{0.4pt}\\[-1mm]
			{\small MNIST}\\[-2mm]
			\rule{\linewidth}{0.4pt}\\[1mm]
			\setlength{\tabcolsep}{0.5pt}
			\begin{tabular}{@{}ccccc@{}}
				\includegraphics[width=0.19\linewidth]{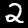} &
				\includegraphics[width=0.19\linewidth]{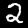} &
				\includegraphics[width=0.19\linewidth]{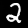} &
				\includegraphics[width=0.19\linewidth]{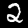} &
				\includegraphics[width=0.19\linewidth]{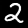} \\[-1mm]
				
				\includegraphics[width=0.19\linewidth]{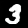} &
				\includegraphics[width=0.19\linewidth]{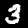} &
				\includegraphics[width=0.19\linewidth]{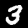} &
				\includegraphics[width=0.19\linewidth]{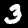} &
				\includegraphics[width=0.19\linewidth]{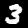} \\[-1mm]
				
				\includegraphics[width=0.19\linewidth]{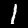} &
				\includegraphics[width=0.19\linewidth]{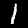} &
				\includegraphics[width=0.19\linewidth]{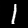} &
				\includegraphics[width=0.19\linewidth]{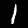} &
				\includegraphics[width=0.19\linewidth]{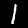}
			\end{tabular}
			
			\rule{\linewidth}{0.4pt}
			
		\end{minipage}
		\hfill
		\begin{minipage}[t]{0.32\textwidth}
			\centering
			
			\rule{\linewidth}{0.4pt}\\[-1mm]
			{\small Fashion MNIST}\\[-2mm]
			\rule{\linewidth}{0.4pt}\\[1mm]
			\setlength{\tabcolsep}{0.5pt}
			\begin{tabular}{@{}ccccc@{}}
				\includegraphics[width=0.19\linewidth]{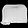} &
				\includegraphics[width=0.19\linewidth]{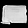} &
				\includegraphics[width=0.19\linewidth]{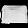} &
				\includegraphics[width=0.19\linewidth]{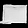} &
				\includegraphics[width=0.19\linewidth]{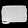} \\[-1mm]
				
				\includegraphics[width=0.19\linewidth]{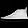} &
				\includegraphics[width=0.19\linewidth]{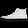} &
				\includegraphics[width=0.19\linewidth]{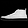} &
				\includegraphics[width=0.19\linewidth]{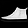} &
				\includegraphics[width=0.19\linewidth]{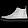} \\[-1mm]
				
				\includegraphics[width=0.19\linewidth]{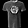} &
				\includegraphics[width=0.19\linewidth]{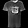} &
				\includegraphics[width=0.19\linewidth]{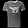} &
				\includegraphics[width=0.19\linewidth]{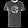} &
				\includegraphics[width=0.19\linewidth]{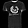}
			\end{tabular}
			
			\rule{\linewidth}{0.4pt}
			
		\end{minipage}
		\caption{
			Qualitative comparison on FFHQ, MNIST, and Fashion-MNIST.
			From left to right, each row presents the query image, the generated image,
			and the Top-3 nearest neighbors retrieved from the latent space.
		}
		\label{fig:qualitative_comparison}
	\end{figure*}

	Example generated images are presented in Fig. \ref{fig:qualitative_comparison} for FFHQ, MNIST, and Fashion-MNIST datasets. The first column from left in each dataset is the chosen query, the second is the generated image, and the last three are nearest neighbors, respectively. Generated images are quite natural and innovative with overall similarity to the query image while they receive special features from neighbors.

\begin{figure*}[t]
		\centering
		
		\begin{minipage}[t]{1\textwidth}
			\centering
			
			\setlength{\tabcolsep}{0.pt}
			\begin{tabular}{@{}cccccccccc@{}}
				\includegraphics[width=0.1\linewidth]{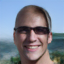} &
				\includegraphics[width=0.1\linewidth]{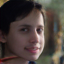} &
				\includegraphics[width=0.1\linewidth]{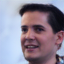} &
				\includegraphics[width=0.1\linewidth]{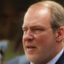} &
				\includegraphics[width=0.1\linewidth]{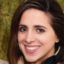} &
				\includegraphics[width=0.1\linewidth]{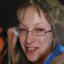} &
				\includegraphics[width=0.1\linewidth]{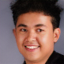} &
				\includegraphics[width=0.1\linewidth]{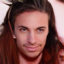} &
				\includegraphics[width=0.1\linewidth]{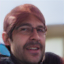} &
				\includegraphics[width=0.1\linewidth]{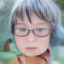} \\[-1mm]
				
				\includegraphics[width=0.1\linewidth]{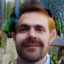} &
				\includegraphics[width=0.1\linewidth]{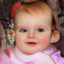} &
				\includegraphics[width=0.1\linewidth]{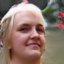} &
				\includegraphics[width=0.1\linewidth]{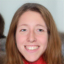} &
				\includegraphics[width=0.1\linewidth]{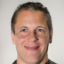} &
				\includegraphics[width=0.1\linewidth]{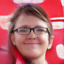} &
				\includegraphics[width=0.1\linewidth]{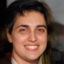} &
				\includegraphics[width=0.1\linewidth]{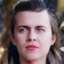} &
				\includegraphics[width=0.1\linewidth]{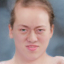} &
				\includegraphics[width=0.1\linewidth]{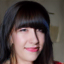} \\[-1mm]
				
				\includegraphics[width=0.1\linewidth]{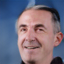} &
				\includegraphics[width=0.1\linewidth]{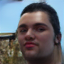} &
				\includegraphics[width=0.1\linewidth]{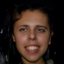} &
				\includegraphics[width=0.1\linewidth]{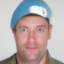} &
				\includegraphics[width=0.1\linewidth]{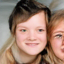} &
				\includegraphics[width=0.1\linewidth]{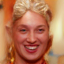} &
				\includegraphics[width=0.1\linewidth]{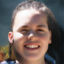} &
				\includegraphics[width=0.1\linewidth]{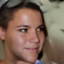} &
				\includegraphics[width=0.1\linewidth]{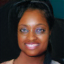} &
				\includegraphics[width=0.1\linewidth]{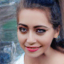} \\[-1mm]
			\end{tabular}

		\end{minipage}
		\caption{Generated samples on FFHQ 64$\times$64}
		\label{fig:ffhq64}
	\end{figure*}

\begin{figure*}[t]
		\centering
		
		\begin{minipage}[t]{1\textwidth}
			\centering

			\setlength{\tabcolsep}{0.pt}
			\begin{tabular}{@{}cccccc@{}}
				\includegraphics[width=0.165\linewidth]{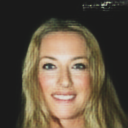} &
				\includegraphics[width=0.165\linewidth]{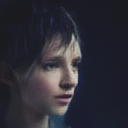} &
				\includegraphics[width=0.165\linewidth]{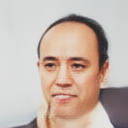} &
				\includegraphics[width=0.165\linewidth]{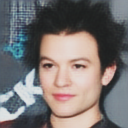} &
				\includegraphics[width=0.165\linewidth]{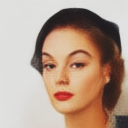} &
				\includegraphics[width=0.165\linewidth]{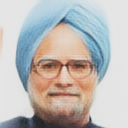} \\[-1mm]
				
				\includegraphics[width=0.165\linewidth]{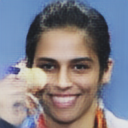} &
				\includegraphics[width=0.165\linewidth]{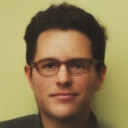} &
				\includegraphics[width=0.165\linewidth]{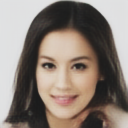} &
				\includegraphics[width=0.165\linewidth]{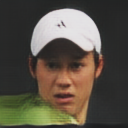} &
				\includegraphics[width=0.165\linewidth]{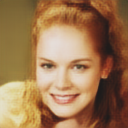} &
				\includegraphics[width=0.165\linewidth]{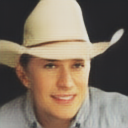} \\[-1mm]
			\end{tabular}

		\end{minipage}
		\caption{Generated samples on CelebA 128$\times$128}
		\label{fig:celeba}
	\end{figure*}

	More generated sample images is shown in Fig. \ref{fig:ffhq64}. It can be seen that most of the images are natural and of good quality. The higher resolution of 128$\times$128 with CelebA dataset is also examined in Fig. \ref{fig:celeba}.

	Baseline diffusion model configurations are listed in Table \ref{tab:baseline_config}. A different setting is used for each dataset used for comparison. Similar settings of the proposed method is also listed in Table \ref{tab:proposed_config}. Table \ref{tab:fid_comparison} compares FID of the proposed method versus the baseline diffusion model. While our proposed model has much less parameters, it achieves better FID in MNIST and comparable FID in Fashio MNIST and CelebA, while the performance is lower in FFHQ dataset. Table \ref{tab:ablation} shows an optimization study which shows that lower $\sigma$ parameter of the attention mechanism will result in better FID metric.

	
	\begin{table}[t]
		\centering
		\caption{Training configurations of the baseline diffusion models.}
		\label{tab:baseline_config}
		\scriptsize
		\setlength{\tabcolsep}{3pt}
		\renewcommand{\arraystretch}{1.05}
		\begin{tabular}{|l|c|c|c|c|}
			\hline
			& MNIST & FashionMNIST & FFHQ & CelebA \\
			\hline
			$f$ & 1 & 1 & 4 & 4 \\
			\hline
			Diffusion steps & 1000 & 1000 & 1000 & 1000 \\
			\hline
			Noise schedule & linear & linear & linear & linear \\
			\hline
			Backbone & CNN & CNN & Transformer & CNN \\
			\hline
			\#Params & 15.7M & 15.7M & 130M & 47M \\
			\hline
			Attention resolution & 7 & 7 & -- & 16, 8 \\
			\hline
			\#Heads & 4 & 4 & 12 & 1 \\
			\hline
			Batch size & 256 & 256 & 256 & 32 \\
			\hline
			Iterations & 30k & 36k & 134k & 51k \\
			\hline
			Learning rate & $10^{-4}$ & $10^{-4}$ & $10^{-4}$ & $10^{-4}$ \\
			\hline
		\end{tabular}
	\end{table}

	
	\begin{table}[t]
		\centering
		\caption{Architecture and training configurations of the proposed latent-space models.}
		\label{tab:proposed_config}
		\scriptsize
		\setlength{\tabcolsep}{1.5pt}
		\renewcommand{\arraystretch}{1.05}
		\begin{tabular}{|l|c|c|c|c|}
			\hline
			& MNIST & FashionMNIST & FFHQ & CelebA \\
			\hline
			$f$ & 4 & 4 & 8 & 8 \\
			\hline
			$z$-shape
			& $3\times7\times7$
			& $3\times7\times7$
			& $16\times8\times8$
			& $4\times16\times16$ \\
			\hline
			Latent Signals
			& 1024
			& 1024
			& 2048
			& - \\
			\hline
			Backbone & CNN & CNN & CNN & CNN \\
			\hline
			\#Params
			& 2.95M
			& 2.95M
			& 64.8M
			& 53M \\
			\hline
			Attention resolution
			& --
			& --
			& \makecell{Encoder: 8\\Decoder: 8, 16}
			& -- \\
			\hline
			\#Heads
			& --
			& --
			& 1
			& -- \\
			\hline
			Batch size
			& 256
			& 256
			& 256
			& 32 \\
			\hline
			Iterations
			& 13.6k
			& 31.3k
			& 44k
			& 51k \\
			\hline
			Learning rate
			& $10^{-4}$
			& $10^{-4}$
			& $10^{-4}$
			& $10^{-4}$ \\
			\hline
		\end{tabular}
	\end{table}

	
	\begin{table}[t]
		\centering
		\caption{Quantitative comparison of the baseline and proposed methods in terms of FID.}
		\label{tab:fid_comparison}
		\small
		\setlength{\tabcolsep}{5pt}
		\renewcommand{\arraystretch}{1.1}
		\begin{tabular}{|l|l|c|}
			\hline
			Dataset & Method & FID \\
			\hline
			\multirow{2}{*}{MNIST}
			& Diffusion & 19.30 \\
			\cline{2-3}
			& Ours & 8.60 \\
			\hline
			\multirow{2}{*}{FMNIST}
			& Diffusion & 13.00 \\
			\cline{2-3}
			& Ours & 14.60 \\
			\hline
			\multirow{2}{*}{FFHQ}
			& Diffusion & 8.60 \\
			\cline{2-3}
			& Ours & 16.58 \\
			\hline
			\multirow{2}{*}{CelebA}
			& Diffusion & 30.49 \\
			\cline{2-3}
			& Ours & 32.15 \\
			\hline
		\end{tabular}
	\end{table}

	
	\begin{table}[t]
		\centering
		\caption{Ablation study of the proposed sampling method under different $\sigma$ and Top-$K$ settings on the FFHQ dataset.}
		\label{tab:ablation}
		\small
		\setlength{\tabcolsep}{5pt}
		\renewcommand{\arraystretch}{1.1}
		\begin{tabular}{|c|c|c|c|}
			\hline
			$\sigma$ & Top-$K$ & Noise scale & FID \\
			\hline
			30    & 3 & 0.01 & 27.43 \\
			\hline
			30    & 5 & 0.01 & 34.33 \\
			\hline
			0.01  & 3 & 0.01 & 16.69 \\
			\hline
			0.001 & 3 & 0.01 & 16.58 \\
			\hline
		\end{tabular}
	\end{table}

\end{document}